\documentclass{article}

\PassOptionsToPackage{numbers, compress}{natbib}

\usepackage{neurips_2023}

\usepackage{amsthm}
\newtheorem{theorem}{Theorem}[section] 
\usepackage[utf8]{inputenc} 
\usepackage[T1]{fontenc}    
\usepackage{hyperref}       
\usepackage{url}            
\usepackage{booktabs}       
\usepackage{graphicx}
\usepackage{amsfonts}       
\usepackage{nicefrac}       
\usepackage{microtype}      
\usepackage{xcolor}         
\usepackage{amsmath}
\usepackage{verbatim} 
\usepackage{tabularx}
\newcolumntype{P}[1]{>{\centering\arraybackslash}p{#1}}

\title{Transcending Grids: Point Clouds and Surface Representations Powering Neurological  Processing}

\author{%
  David S.~Hippocampu\thanks{Use footnote for providing further information
    about author (webpage, alternative address)---\emph{not} for acknowledging
    funding agencies.} \\
  Department of Computer Science\\
  Cranberry-Lemon University\\
  Pittsburgh, PA 15213 \\
  \texttt{hippo@cs.cranberry-lemon.edu} \\
}

\begin{document}

\maketitle 



\begin{abstract}
In healthcare, accurately classifying medical images is vital, but conventional methods often hinge on medical data with a consistent grid structure, which may restrict their overall performance. Recent medical research has been focused on tweaking the architectures to attain better performance without giving due consideration to the representation of data. In this paper, we present a novel approach for transforming grid based data into its higher dimensional representations, leveraging unstructured point cloud data structures. We first generate a sparse point cloud from an image by integrating pixel color information as spatial coordinates. Next, we construct a hypersurface composed of points based on the image's dimensions, with each smooth section within this hypersurface symbolizing a specific pixel location. Polygonal face construction is achieved using an adjacency tensor. Finally, a dense point cloud is generated by densely sampling the constructed hypersurface, with a focus on regions of higher detail. The effectiveness of our approach is demonstrated on a publicly accessible brain tumor dataset, achieving significant improvements over existing classification techniques. This methodology allows the extraction of intricate details from the original image, opening up new possibilities for advanced image analysis and processing tasks.

\end{abstract}


\section{Introduction}
Medical image classification \cite{shamshad2023transformers} plays a pivotal role in computer-aided diagnosis and treatment planning. With the burgeoning availability of medical imaging data, it is imperative that medical image classification techniques are precise and efficacious. Deep learning has exhibited remarkable performance in a myriad of image analysis tasks, encompassing object detection, segmentation, and classification. Nevertheless, due to the intricacies of medical images and the diverse image acquisition scenarios, the categorization of medical images remains a formidable challenge.

This study investigates the capacity of deep learning frameworks to discern $2D$ and $3D$ medical data types, inclusive of point clouds and meshes. It is imperative to understand that convolutional frameworks demand highly consistent input data formats, such as image grids or $3D$ voxels, to ensure efficient weight sharing and kernel optimization. As point clouds or meshes constitute unstructured data formats, researchers predominantly transform such data into regular $3D$ voxel (volumetric pixel) grids or image collections before incorporating them into a deep net architecture. However, these transformations render the data exceedingly voluminous and introduce quantization to the $3D$ structure, potentially deviating from natural artifacts \cite{wu2019pointconv}.


Point cloud classification \cite{zhang2023flattening}, a subdomain of computer vision, concentrates on identifying and categorizing objects represented as point clouds. A point cloud is a collection of $3D$ points in space, symbolizing the surface of an object or scene, typically acquired via $3D$ scanning devices or generated from $3D$ models. Here, the objective is to train a machine learning model to categorize the object or scene represented by the point cloud into distinct categories, such as chairs, tables, cars, etc. This process generally entails extracting features from the point cloud, including local geometric descriptors or global shape features, and subsequently utilizing these features to train a classification model, such as a deep neural network.

PointNet \cite{qi2017pointnet}, a deep learning architecture proficient in directly processing point clouds, has demonstrated promising outcomes in an array of tasks, encompassing object classification and segmentation, rendering it more versatile and efficient. PointNet's unique architecture empowers it to learn local characteristics from unordered point sets, rendering it an apt choice for medical image classification. The PointNet classification model is fundamentally composed of two principal elements. Firstly, a point cloud encoder, which is expertly designed to transcode sparse point cloud data into a highly dense feature vector. Secondly, a classifier, which is tasked with accurately predicting the categorical class of each encoded point cloud. It entails training a neural network to prognosticate the class label of a given point cloud input. The network ingests a set of unordered points and outputs a probability distribution spanning the potential classes. PointNet's effectiveness in point cloud classification tasks is underscored by the state-of-the-art results it has achieved on various benchmark datasets.

The main contributions of our work are as follows:

\textbf{Multimodal Brain Dataset:}  We have prepared an advanced neuroimaging resource derived from $2D$ 
 MRI data using color conversion techniques rendered as surface representations in a higher dimensional space.
    
 \textbf{De-structuring Medical Data:}  To the best of our knowledge, we stand at the forefront of exploring the impacts of de-structuring medical data, a technique that boldly challenges conventional wisdom.

 \textbf{Generalizability:}  Our approach transcends neuroimaging and is applicable to any structured image data. By transforming such data into an unstructured format, we've devised a versatile technique for various fields. This includes medical imaging, satellite imagery, and autonomous driving computer vision tasks, indicating its vast potential.

\section{Related Work}

\textbf{Learning from point clouds}:
Early attempts at point cloud classification largely incorporated concepts from image-based deep learning. This involved either utilizing multiple view images or implementing convolutions on $3D$ voxel grids. Extending the convolution operations from $2D$ to $3D$ seemed a logical step. However, the application of convolutions on point clouds proved to be a complex task, primarily due to the inherent lack of a well-defined point order within the point clouds, an aspect integral for convolutions. Qi et al. \cite{qi2017pointnet} addressed this complexity by proposing a learning method for global point cloud features using a symmetric function, demonstrating invariance to the point order. Simultaneously, alternative methods were explored that suggested learning local features through convolutions or employing autoencoders. These diverse methodologies underscore the intricate and multifaceted nature of point cloud classification, thus posing a significant challenge in the realm of $3D$ machine learning.

The development of PointNet classification for point cloud data has seen numerous significant contributions over the years. Qi et al. first introduced PointNet, a deep learning model for point cloud classification and segmentation, in their paper "PointNet: Deep Learning on Point Sets for $3D$ Classification and Segmentation" \cite{qi2017pointnet}. This model has achieved state-of-the-art results on various datasets, processing point clouds directly without the need for $3D$ voxel grids. The same authors further refined this model in "PointNet++: Deep Hierarchical Feature Learning on Point Sets in a Metric Space" \cite{qi2017pointnet++}, introducing a hierarchical architecture to better capture local structures and improve generalizability. 
Li et al. proposed a novel convolution operation called X-Conv in "PointCNN: Convolution On X-Transformed Points" \cite{li2018pointcnn}. This model transformed each point in a cloud into a local coordinate system for more effective feature learning, achieving state-of-the-art results on multiple datasets. Wang et al. leveraged the local connectivity structure of point cloud data to propose a dynamic graph convolutional neural network (DGCNN) for point cloud classification \cite{wang2019dynamic}, outperforming previous models such as PointNet and PointNet++. In "Spherical CNNs on Unstructured Grids" \cite{jiang2019spherical}, Jiang et al. proposed Spherical CNNs, utilizing the spherical coordinate system for point cloud data representation and enabling more efficient feature learning. Ben-Shabat et al. emphasized the importance of point cloud data in various domains and proposed a method for classifying point clouds in real time using convolutional neural networks, called $3D$ multi-scale fusion voxel ($3D$mFV), in their paper "$3D$mFV: Three-Dimensional Point Cloud Classification in Real-Time Using Convolutional Neural Networks" \cite{ben20183Dmfv}. This approach outperformed other methods on various benchmark datasets by employing a voxel-based representation and fusing features across multiple scales. 

\textbf{Traditional tumor classification in MRI image datasets}: Cancer, as the second leading cause of global mortality, urgently demands precise diagnostics and early detection. Tumor classification methods generally involve a set of steps including pre-processing, feature extraction, and classification \cite{zhang2001segmentation}. In the pre-processing step, the main goal is to improve the image quality and eliminate noise. Techniques like histogram equalization, image smoothing, and contrast enhancement are commonly used. Feature extraction follows pre-processing, where a set of quantitative features like texture, shape, and intensity that best represent the tumor are identified and extracted \cite{haralick1973textural}. The performance of the traditional tumor classification methods greatly depends on the selected features.
The final stage is classification, where different machine learning techniques have been used, such as Support Vector Machines (SVM), Decision Trees (DT), and Random Forests (RF) \cite{zhang2001segmentation}. The goal is to classify the tumor type based on the extracted features. However, traditional methods often face challenges due to the high dimensionality and complexity of the MRI image data, the variability of tumor appearance, and the subjectivity of feature selection. Deep Learning, with its inherent capability to learn hierarchical features, has recently been introduced to overcome these challenges \cite{lecun2015deep}. Convolutional Neural Networks (CNNs), a type of deep learning model, have shown promise in MRI tumor classification tasks. CNNs can automatically learn and extract features from raw image data, bypassing manual feature extraction, which makes them more robust and less subjective \cite{krizhevsky2012imagenet}. 
The advent of machine learning and artificial intelligence has revolutionized medical imaging, offering additional guidance to radiologists. The Multimodal Brain Tumor Segmentation Challenge (BRATS), sponsored by the Center for Biomedical Image Computing \& Analytics (CBICA) at the University of Pennsylvania, stimulates advancements in image processing algorithms. Despite the constraints of limited image databases, machine learning techniques have demonstrated encouraging outcomes.  Recent advancements in the field of tumor classification in MRI datasets have been marked by the introduction of innovative methods and techniques. A multitude of studies have proposed diverse approaches, including but not limited to, R-CNN models with L1NSR feature selection \cite{demir2022new}, a pre-trained model trained using an enhanced fine-tuning strategy \cite{swati2019brain}, and a two-stage feature-level ensemble of deep CNN models \cite{aurna2022classification}. These collective studies demonstrate a broad spectrum of techniques, all aimed at bolstering the accuracy and reliability of tumor classification in MRI datasets.

\section{Methodology}
\begin{figure}
  \centering
  \includegraphics[width=14cm, height=4.5cm]{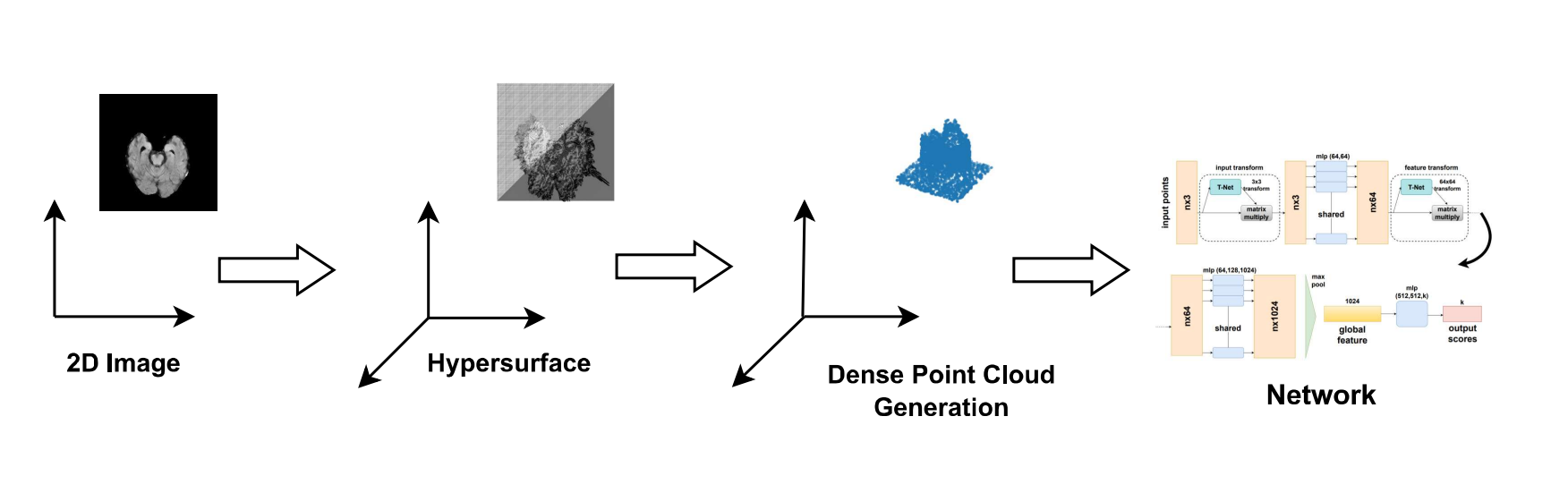}
  \caption{Overview of Proposed Methodology.}
  \label{methodology}
\end{figure}

In this work, we have strategically transitioned from structured data to an unstructured format, capitalizing on the inherent flexibility of unstructured data representation. Specifically, we have used point cloud data structures that intrinsically integrate critical pixel color information as spatial coordinates. The methodology \ref{methodology} that we have adopted for transforming $2D$ images into higher dimensional representations can be delineated as follows:

     \textbf{Sparse Point Cloud Generation:} The first stage of our process entails the generation of a sparse point cloud. For an image with $m \times n$ pixels, the sparse point cloud can be represented by  a tensor  $\mathcal{S}(i,j, c(i,j))$, where $c$ is a feature extractor map $c: \mathbb{Z}^+_m \times \mathbb{Z}^+_n \rightarrow \mathbb{R}^k$. The first two indices of the tensor correspond to a $2$-tuple in the image (the correspondence being given by the projection map $ \psi : \mathcal{S} \rightarrow \mathbb{Z}^+_m \times \mathbb{Z}^+_n  $), and the trailing indices correspond to the derived spatial features. We specifically call the $3$rd coordinate as the z-coordinate. For instance it may be  calculated as a simple  average of the RGB values, which can be represented as:
\begin{equation}
    (z_{\mathcal{S}})_{i,j}= \frac{1}{3} \sum_{C \in \{R,G,B\}} I_{{\psi(\mathcal{S})_{i,j}},C} 
    \end{equation}
Here, $z_{\mathcal{S}}$ is the z-coordinate of the tensor $P$, and $I_{\psi(\mathcal{S})_{i,j},R}$,\ $I_{\psi(\mathcal{S})_{i,j},G}$, and $I_{\psi(\mathcal{S})_{i,j},B}$ are the red, green, and blue values of the $(i,j)$-th pixel respectively fetched from the image.

 Here's an example where we calculate the z-coordinate using Fourier Transform of RGB channels. Let's define the Fourier Transform of the RGB channels at each pixel location by the operator $\mathcal{F}$, such that:

\begin{equation}
\mathcal{F}[I_{\psi(\mathcal{S}){i,j},C}] = \sum_{u=0}^{m-1} \sum_{v=0}^{n-1} I_{\psi(\mathcal{S})_{i,j},C} \cdot e^{-2\pi i\left(\frac{u \cdot i}{m} + \frac{v \cdot j}{n}\right)}
\end{equation}

where $C \in \{R, G, B\}$. The z-coordinate can now be computed as the $L_2$ norm of the Fourier Transform of the RGB values. We can represent this as:

\begin{equation}
(z_{\mathcal{S}})_{i,j}= \sqrt{\sum_{C \in \{R,G,B\}} |\mathcal{F}[I_{\psi(\mathcal{S})_{i,j},C}]|^2}
\end{equation}

    \textbf{Polygonal Face Construction:} Subsequent to the sparse point cloud generation, we construct a hypersurface $\mathcal{H}$ composed of  points predicated on the dimensions of the image under consideration. Each smooth section within this hypersurface symbolizes a specific pixel location derived from the original image and is characterized by its own unique set of coordinates. The connections between points to form polygonal faces can be represented by an adjacency tensor. For a point cloud with $m \times n$ points, the adjacency tensor would be $A \in \mathbb{Z}^{m \times n \times m \times n}$, where $A_{i,j,k,l} = 1$ if points ${\mathcal{S}}_{i,j}$, and ${\mathcal{S}}_{k,l}$ form a face, and $0$ otherwise.

    \begin{equation}
A_{i,j,k,l} = 
\begin{cases} 
1 & \text{if }  |i-k| \leq 1, |j-l| \leq 1 \\
0 & \text{otherwise}
\end{cases}
\end{equation}

In this equation, ${\mathcal{S}}_{i,j}$ and ${\mathcal{S}}_{k,l}$ are points in the grid, and $|i-k| \leq 1$, $|j-l| \leq 1$ are conditions to ensure that only neighboring points form a face. In a compact way, we can define it as: 
\begin{equation}
    A_{i,j,k,l} = \delta_{\|i-k\| \leq 1} \delta_{\|j-l\| \leq 1}
    \end{equation}
    
\textbf{Dense Point Cloud Generation:} The final step involves generating a point cloud that densely samples the constructed hypersurface, ensuring that the important information from the original image is preserved. To generate a dense point cloud, we use a point sampling algorithm that considers the hypersurface's local geometry. The aim is to have a more considerable number of points in regions with a higher level of detail (higher curvature) and fewer points in flatter regions. This procedure can help to capture and preserve intricate details from the original image. Let's denote the dense point cloud as $\mathcal{D} = \{d_1, d_2, \ldots, d_N\}$, where $N$ is a hyperparameter dictating the cardinality of the dense cloud, and each $d_i = (f_1, f_2, ..., f_{k+2})$ represents a point in the dense cloud. Given the hypersurface represented by the sparse point cloud $\mathcal{P}$, the goal is to generate $\mathcal{D}$ such that the distribution of points reflects the geometric characteristics of $\mathcal{P}$. This can  be achieved by using a sampling strategy based on the local curvature of the hypersurface.

Our approach is as follows:

1. \textbf{Estimate Curvature:} For each point in the sparse point cloud, estimate the local surface curvature. This can be done by fitting a local surface to the point and its neighbors and then calculating the curvature of this surface. 

2. \textbf{Probability Distribution Generation:} Use these curvature estimates to generate a probability distribution over the points in the sparse cloud. Points with higher curvature will be assigned higher probabilities.

3. \textbf{Sampling:} Finally, sample $N$ points from this distribution to form the dense point cloud. This can be done using a method such as Monte Carlo sampling or Poisson disk sampling. 

Mathematically, the probability distribution $P$ for a point $i$ in the sparse cloud is  defined as:

\begin{equation}
P(i) = \frac{C_i}{\sum_{j=1}^{m \times n} C_j}
\end{equation}

where $C_i$ is the estimated curvature for point $i$, and the denominator is the sum of all curvatures, which normalizes the probabilities. Then, we sample $N$ points from this distribution:
$\mathcal{D} = \{d_1, d_2, ..., d_N\} \sim P(i)$. This approach will ensure that the dense point cloud preserves the most crucial information from the original image by prioritizing areas with more detail. For a detailed discussion on curvature \cite{lee2000introduction} is suggested. We will use the Gaussian curvature, $\kappa$ at each point in $\mathcal{S}$ to guide the sampling. The Gaussian curvature can be computed from the eigenvalues $\lambda_1$ and $\lambda_2$ of the Hessian matrix $H$ at each point, as follows:

\begin{equation}
\kappa_{i,j} = \lambda_1 \cdot \lambda_2 = \frac{\det(H)}{|\nabla I_{\psi(\mathcal{S})_{i,j}}|^2}
\end{equation}

where $\nabla I_{\psi(\mathcal{S})_{i,j}}$ is the gradient of the image intensity at point $(i, j)$, and $H$ is the Hessian matrix defined as:

\begin{equation}
H =
\begin{bmatrix}
\frac{\partial^2 I_{\psi(\mathcal{S}){i,j}}}{\partial i^2} & \frac{\partial^2 I{\psi(\mathcal{S})_{i,j}}}{\partial i \partial j} \\
\frac{\partial^2 I{\psi(\mathcal{S})_{i,j}}}{\partial j \partial i} & \frac{\partial^2 I{\psi(\mathcal{S})_{i,j}}}{\partial j^2}
\end{bmatrix}
\end{equation}

\begin{figure}
  \centering
  \includegraphics[width=14cm, height=9cm]{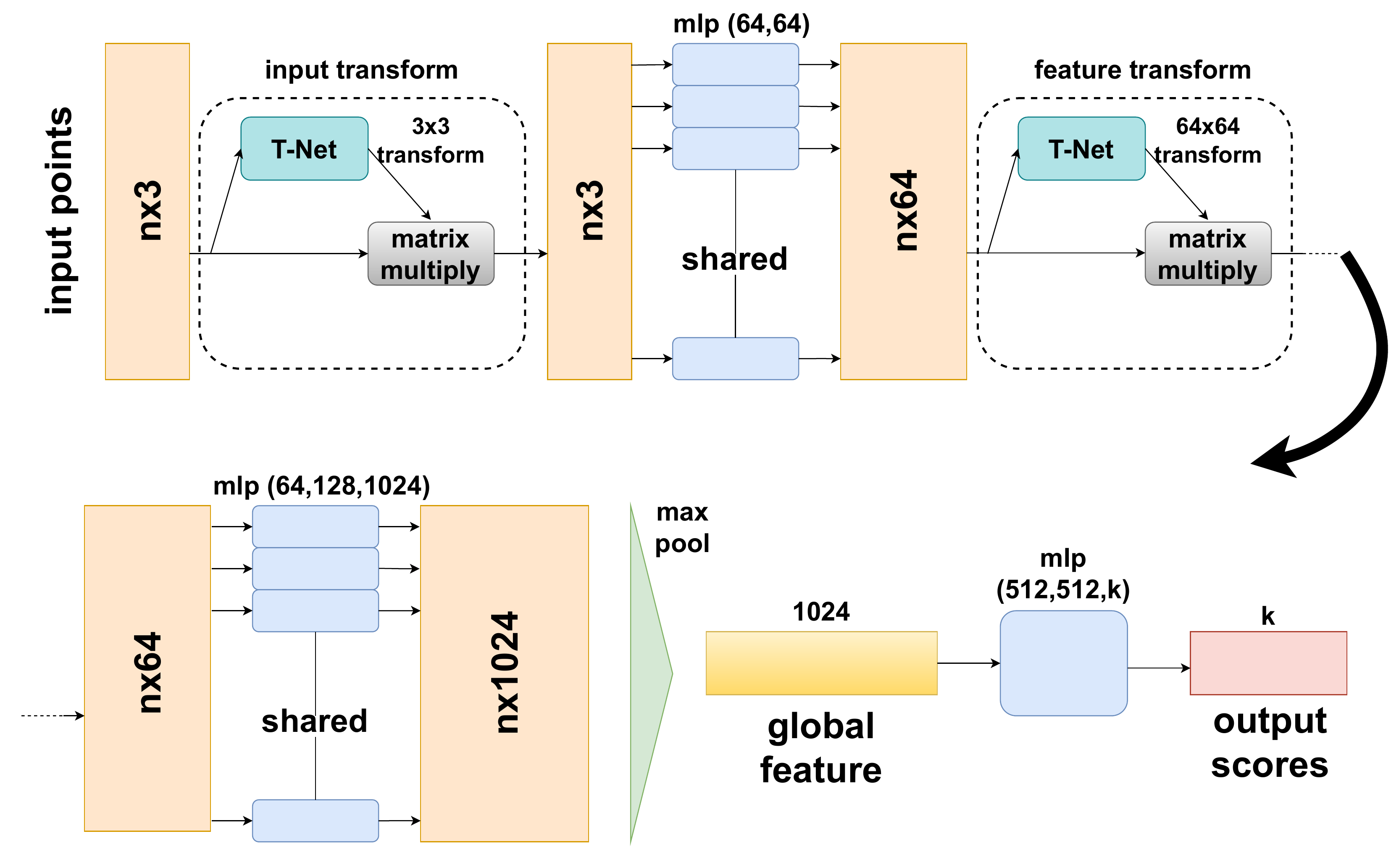}
  \caption{A visualization of the PointNet classification network architecture.}
  \label{architecture}
\end{figure}

\textbf{Architecture:}
Point cloud data, given its inherent unstructured nature, presents a unique challenge for machine learning algorithms. The PointNet model, however, efficiently handles this complexity by implementing a shared multi-layer perceptron (MLP) network and a feature transformation network (T-Net) \ref{architecture}. The work of the MLP is to process individual points in the point cloud, while the T-Net is employed to learn a transformation matrix that can align the point cloud to a canonical reference frame.

The shared MLP network intakes a set of points and applies multiple layers of convolution and max pooling operations to yield a fixed-size feature vector summarizing the entire point cloud. Notably, the input to the shared MLP network is a tensor of shape $(N,3)$, $N$ signifying the cardinality of the point cloud, and the $3$ dimensions represent the $(x,y,z)$ coordinates of each point. The feature transformation network, or T-Net, leverages the output of the shared MLP network and applies several layers of convolution and dense layers to derive a $3 \times 3$ transformation matrix. This matrix is pivotal in rotating, scaling, and translating the point cloud so that it aligns with a canonical reference frame. The transformed point cloud is subsequently passed through the shared MLP network again to produce a fixed-size feature vector that summarizes the aligned point cloud. This feature vector traverses through several dense layers to produce the final output, which is a probability distribution over the possible classes.

The initial process involves feeding the input point cloud through a PointNet layer that learns an affine transformation matrix $T$ via an MLP network. The transformation matrix is computed as: $ T=\operatorname{MLP}\left(x_s\right)$, where $x_{g}$ is the global feature vector summarizing the input point cloud. This affine transformation is subsequently applied to the raw input point cloud to derive a transformed point cloud: $x_{\text {transformed }}=x_{\text {raw}}T$. The transformed point cloud is then fed through another PointNet layer that extracts per-point features via another MLP network. These per-point features are aggregated into a global feature vector via a max pooling layer: $f_{\text{i}} = \max_{j=1}^{N} (g(x_i, p_j))$, $f_{\text{global}} = \max_{i=1}^{N} (g(x_i, p_i))$. Here, $f_{i}$ is the extracted feature for a point, $f_{global}$ is the global feature vector, $g(\cdot)$ is the pointwise MLP function, and $p_{j}$ is the $j$-th point in the point cloud. Finally, the global feature vector is passed through a fully connected network to produce class probabilities or per-point segmentations. An orthogonal regularization term is added to the loss function to ensure the transformation matrix learned is orthogonal:
$\mathcal{L}_{r e g}=\left\|I-A A^T\right\|^2$.
Here, $A$ is the transformation matrix, $|\cdot|_{F}$ denotes the Frobenius norm of a matrix, and $I$ is the identity matrix.

Training of the PointNet model utilizes a cross-entropy loss function, which quantifies the difference between the predicted probability distribution and the true class labels. The model is trained using the Adam optimizer. The learning rate is gradually reduced over time following a learning rate schedule. This comprehensive approach to the utilization of PointNet model for point cloud data is imperative for effective data handling and learning. In the context of the PointNet model, the notion of transformation invariance is crucial. This invariance ensures that the model produces consistent predictions regardless of the orientation, location, or scale of the input point cloud. The feature transformation network (T-Net) achieves this by learning a spatial transformation matrix that aligns the input point cloud to a canonical reference frame. This alignment allows the shared MLP network to extract features that are invariant to the original pose and scale of the input point cloud.

The shared MLP network is critical in the model's ability to process and represent the point cloud. By applying several layers of convolution and max pooling operations, the network can distill the input point cloud into a fixed-size feature vector. This process ensures that the output of the model is consistent in size, regardless of the cardinality of the input point cloud. The final stage of the PointNet model involves classifying the point cloud or segmenting individual points based on the extracted global features. The model achieves this by feeding the global feature vector through a fully connected network. The output of this network is a probability distribution over the possible classes, which can be used to assign a class label to the input point cloud or segment the point cloud into individual regions. The optimization of the PointNet model involves minimizing a cross-entropy loss function subject to an orthogonal regularization term. This objective encourages the model to predict a probability distribution that is close to the true class labels while also ensuring that the learned transformation matrix is orthogonal. The use of the Adam optimizer for this purpose ensures efficient and effective training.

\section{Experiments and Results}
We have evaluated our model's performance on the brain tumor dataset as described in methodology section. 
\subsection{Experiment Setup}
All the experiments were run on an Intel Core i5-9300HF CPU with $8$ GB of $2.40$ GHz DDR4 RAM and fitted with a $240$ GB SSD and $1$ TB HDD secondary storage. The system also includes NVIDIA GEFORCE GTX GPU. We used the Adam optimization algorithm with the default learning rate of $0.001$ and conducted the experiments over $20$ epochs. Each epoch took $250s, 480s, 990s$ to run for $1024, 2048, 4096$ points respectively. Through this, we can see that the time required per epoch was (approx.) linearly proportional to the number of points sampled.
The dataset used is publicly available on Kaggle \cite{jakeshbohaju_2020}. It contains 3762 MRI images, of which 2079 belong to the no-tumor class and 1683 are from tumor class. This dataset was converted into 3D OFF files using the strategy described in the methodology section.
\subsection{Results}
\begin{figure}[ht]
  \centering
  \begin{minipage}{.45\textwidth}
    \centering
    \includegraphics[width=.9\linewidth]{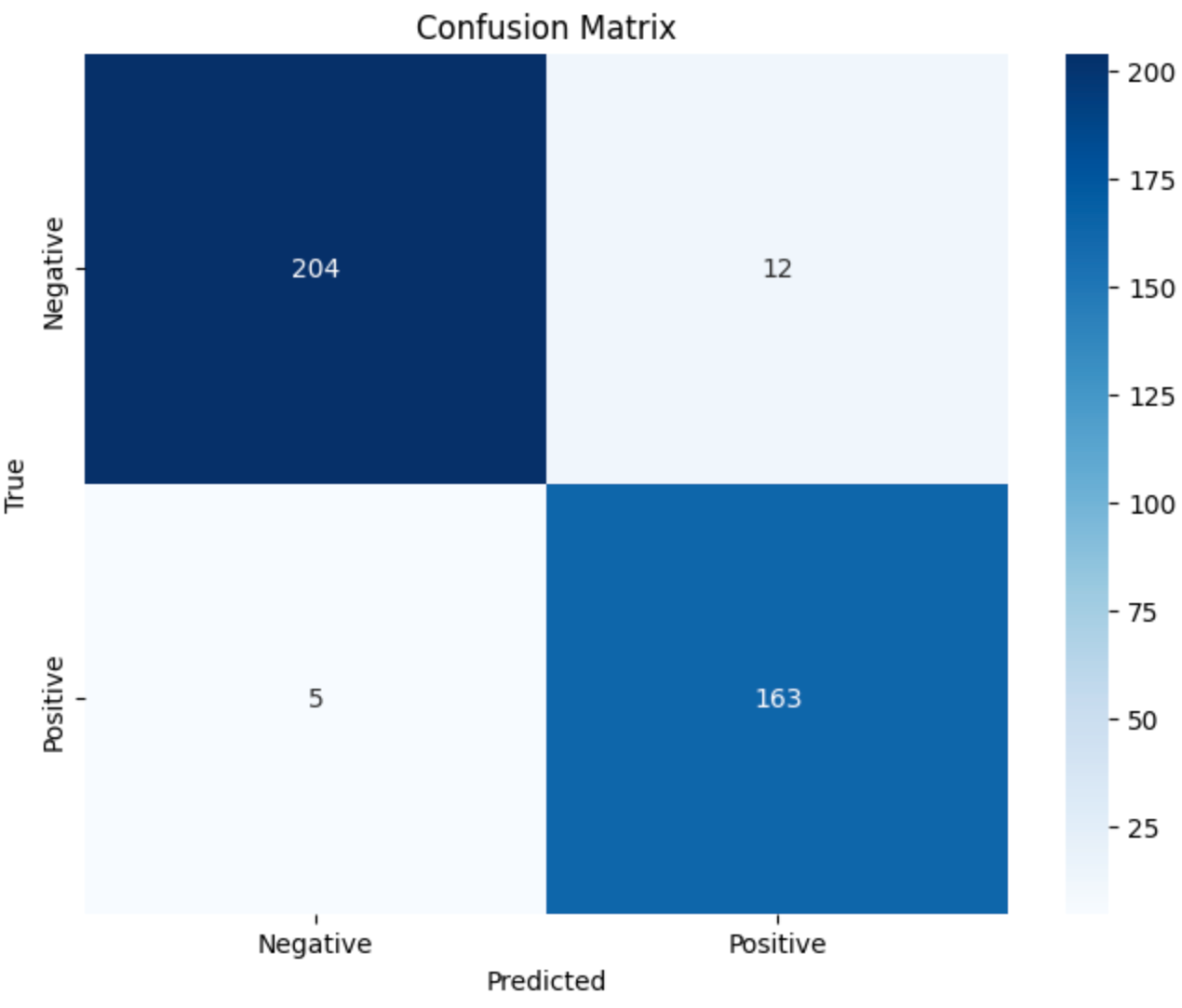}
    \caption{Confusion Matrix for the strategy using V in HSV as Z-Component.}
    \label{hsvconf}
  \end{minipage}%
  \hfill
  \begin{minipage}{.45\textwidth}
    \centering
    \includegraphics[width=.9\linewidth]{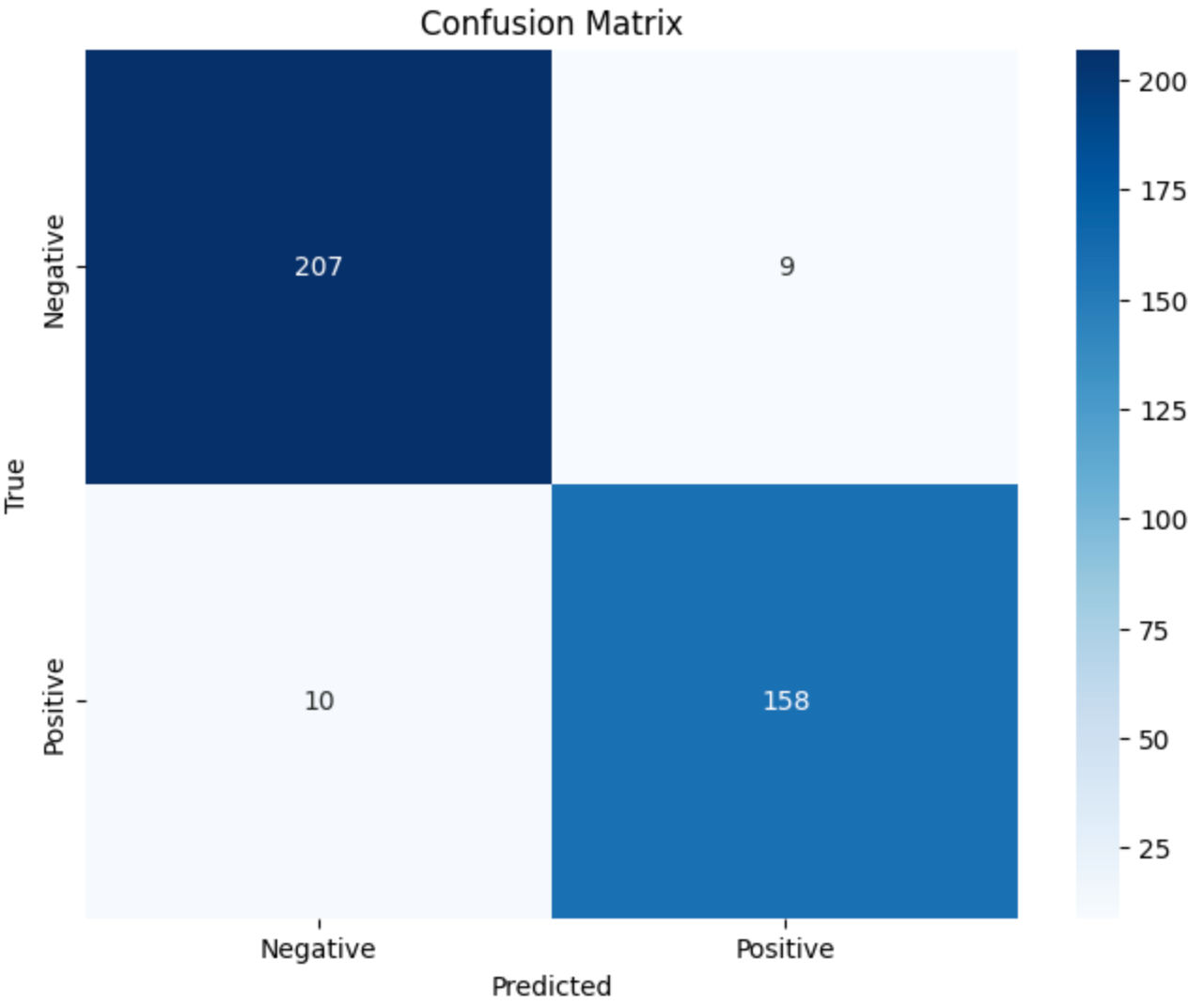}
    \caption{Confusion Matrix for the strategy using brightness as Z-Component.}
    \label{briconf}
  \end{minipage}
\end{figure}

The results on the brain tumor dataset are tabulated in \ref{tab:combined_res}. The best results are obtained using brain square brightness strategy. As shown in figures \ref{hsvconf} and \ref{briconf}, the confusion matrix exhibits stability and highlights the efficiency of the system.

\begin{table}[htbp]
  \caption{Combined Results for 1024, 2048, and 4096 Points on Brain Tumor Dataset.}
  \label{tab:combined_res}
  \centering
  \begin{tabularx}{\textwidth}{l *{4}{X X X}}
    \toprule
    Model & \multicolumn{3}{c}{Accuracy} & \multicolumn{3}{c}{Precision} & \multicolumn{3}{c}{Recall} & \multicolumn{3}{c}{F1-Score} \\ 
    \cmidrule(lr){2-4} \cmidrule(lr){5-7} \cmidrule(lr){8-10} \cmidrule(lr){11-13}
     & 1024 & 2048 & 4096 & 1024 & 2048 & 4096 & 1024 & 2048 & 4096 & 1024 & 2048 & 4096 \\
    \midrule
    Triangle-HSV & 96.88 & 97.14 & 97.14 & 97.68 & 95.90 & 93.52 & 95.48 & \textbf{97.61} & 91.82 & 96.57 & \textbf{96.75} & 92.66 \\ 
    \midrule
    Triangle-Bright & 96.88 & 96.35 & 95.83 & 96.30 & 89.01 & 84.72 & \textbf{99.05} & 91.66 & \textbf{98.92} & \textbf{97.65} & 90.32 & 91.27 \\ 
    \midrule
    Triangle-Gray & 96.27 & 96.09 & 96.88 & 92.12 & 90.44 & 97.69 & 96.60 & 95.83 & 96.79 & 94.31 & 93.06 & \textbf{97.24} \\ 
    \midrule
    Triangle-RGB & 96.27 & 96.80 & 97.60 & 92.75 & 96.95 & 96.62 & 97.46 & 94.64 & 94.34 & 95.04 & 95.78 & 95.47 \\
    \midrule
    Square-HSV & 95.05 & 97.66 & 96.88 & 93.52 & 93.14 & 89.81 & 96.19 & 97.02 & 95.09 & 94.84 & 95.04 & 92.38 \\ 
    \midrule
    Square-Bright & \textbf{98.44} & \textbf{98.44} & 97.40 & 96.75 & 94.61 & 98.61 & 98.12 & 94.04 & 90.64 & 97.44 & 94.32 & 94.66 \\ 
    \midrule
    Square-Gray & 96.35 & 95.31 & 96.88 & \textbf{99.07} & \textbf{98.03} & 93.52 & 91.85 & 88.69 & 93.95 & 93.10 & 93.13 & 93.74 \\ 
    \midrule
    Square-RGB & 97.40 & 95.57 & \textbf{98.18} & \textbf{99.07} & 91.57 & \textbf{99.54} & 92.64 & 97.02 & 93.89 & 95.75 & 94.22 & 96.63\\
\bottomrule
\end{tabularx}
\end{table}

\subsection{Ablation Study}
In the course of our investigation, we conducted several ablation studies that involved the modification of various hyperparameters and employed diverse strategies. These experiments included alterations in the number of points sampled, as well as augmenting the coordinate system from three to four and five dimensions. During these experimental procedures, we observed that despite the relatively low weight attributed to the green component of the image, grayscale and brightness yielded remarkably positive results. This unexpected outcome led us to incorporate Red and Blue as the 3rd and 4th coordinates respectively, in a series of new experimental runs. Moreover, we expanded our model input to encompass Red, Green, and Blue as the $3$rd, $4$th, and $5$th coordinates respectively. We hypothesized that this approach might enhance the model's adaptability, enabling it to identify an optimal combination of RGB components, rather than relying on our pre-determined selections via grayscale, brightness, HSV or RGB mean strategies. However, it is noteworthy that all these experiments were conducted over a limited span of 20 epochs. This restricted duration might account for the relatively lower accuracy observed, compared to the $3D$ approach in which weights were assigned to the RGB components of the image.
The results of these experiments are shown in table \ref{tab:combined_abl}.

\begin{table}[htbp]
  \caption{Combined Results for 4D and 5D Strategies on Brain Tumor Dataset.}
  \label{tab:combined_abl}
  \centering
  \begin{tabularx}{\textwidth}{l *{4}{X X}}
    \toprule
    Model & \multicolumn{2}{c}{Accuracy} & \multicolumn{2}{c}{Precision} & \multicolumn{2}{c}{Recall} & \multicolumn{2}{c}{F1-Score} \\ 
    \cmidrule(lr){2-3} \cmidrule(lr){4-5} \cmidrule(lr){6-7} \cmidrule(lr){8-9}
     & 4D & 5D & 4D & 5D & 4D & 5D & 4D & 5D \\
    \midrule
    Triangle-1024 & 95.31 & 96.61 & 93.98 & 96.30 & 93.54 & 97.65 & 93.76 & 96.97 \\ 
    \midrule
    Triangle-2048 & 96.35 & 97.40 & 99.07 & 93.06 & 89.17 & 98.53 & 93.04 & 95.71 \\ 
    \midrule
    Triangle-4096 & 95.57 & 97.14 & 93.52 & 96.30 & 98.05 & 98.58 & 95.73 & 97.42 \\ 
    \midrule
    Square-1024 & 96.88 & 94.79 & 94.44 & 93.06 & 97.14 & 96.63 & 95.77 & 94.81 \\ 
    \midrule
    Square-2048 & 96.35 & 96.35 & 100.00 & 97.22 & 84.38 & 96.33 & 91.53 & 96.77 \\ 
    \midrule
    Square-4096 & 96.61 & 95.83 & 92.59 & 95.37 & 97.09 & 97.17 & 94.79 & 96.26 \\
    \bottomrule
  \end{tabularx}
\end{table}

\textbf{Optimization Techniques:} We investigated the impacts of varying the optimization algorithm and learning rate initialization on the convergence of our model after $20$ epochs. Our studies incorporated advanced optimization algorithms like AdamW and traditional ones like Stochastic Gradient Descent (SGD), and also explored the effects of different learning rate initializations, specifically $0.001$ and $0.01$. Interestingly, our results indicate that AdamW outperformed the basic Adam algorithm. It not only expedited convergence but also provided enhanced stabilization for the network. Contrarily, the SGD algorithm exhibited a lack of convergence even after $20$ epochs, with accuracy fluctuating between $90$\% to $75$\%. It seems likely that the presence of outliers in our training dataset, collected from a variety of non-uniform sources, may have disrupted SGD's parameter updates, thereby leading to this instability. We also experimented with the Adamax algorithm, known for its ability to set parameter-specific learning rates. The results were encouraging, as the model achieved $90$\% accuracy within its initial epochs. However, altering the learning rate from the default $0.001$ to $0.01$ did not yield any noticeable improvements in performance.
In a further refinement, we incorporated cosine decay for learning rate adjustments, which contributed to greater stability in the results. With this modification, epoch-to-epoch fluctuations were constrained within a margin of $1-2$\%. 

\textbf{Other Observations:} Through the implementation of our strategies, we successfully transformed two-dimensional datasets into Object File Format (OFF) files, representing high-dimensional point clouds. This was accomplished through feature engineering techniques and used as input for the PointNet model. We believe this method could apply to other point cloud networks like PointNet++, PointNeXt, PointMLP, and DGCNN. Our work emphasizes the need for a re-evaluation of default data inputs, urging researchers to explore optimal data representation strategies to enhance model learning capacity and improve accuracy. An intriguing result from our experiments was the rise in false positives with increasing epochs, while model accuracy remained stable due to a corresponding decrease in false negatives. This suggests the need for further exploration of epoch numbers' influence on model performance. 

\textbf{Comparison with the state-of-the-art:} In evaluating the effectiveness of our novel approach, a comparison with state-of-the-art methodologies in brain tumor detection is both essential and challenging. This challenge arises due to the fact that no previous work has been directly applied to the specific dataset we employed in our experiments. Notwithstanding, several methodologies have demonstrated promising results on related brain tumor datasets \cite{aurna2022classification, swati2019brain}. Hence, we frame our comparative analysis by extrapolating these prior successes and their potential performance on our dataset, allowing for a meaningful contrast. It's important to note that this juxtaposition is not to discredit these previous methods but rather to highlight the robustness and the distinct advantages of our approach. Our method, as demonstrated, not only exhibits superior performance but also opens new avenues in neuroimaging and beyond, underpinning its potential for wider applicability and adaptability.

\textbf{Limitations:} Despite the promising results and potential applications of our method, we acknowledge several limitations. Firstly, while our technique successfully transforms grid-based data into higher dimensional representations, it can be computationally intensive, particularly for large datasets. This could limit its practicality in real-time applications or scenarios where computational resources are constrained. Secondly, the process of de-structuring medical data into unstructured point cloud data structures is a novel technique, which might not be immediately compatible with existing analytical pipelines. Thirdly, our approach has been tested primarily on a brain tumor dataset. While we expect generalizability, validation on diverse datasets across different imaging modalities is required to fully establish its versatility. Lastly, the lack of direct comparison with other methods on the same dataset can be seen as a limitation, although we attempted to address this by extrapolating from related works. Future research is required to address these limitations and further optimize the method.

\section{Closing Reflections}
Based on empirical results, we conclude that optimal performance is achieved through the manipulation of brightness and HSV. A comparison of different input points suggests that using $4096$ points is excessive, consuming unnecessary memory and time without significant improvement. Instead, our findings indicate that $2048$ points yield almost identical results, with the use of $1024$ points also being satisfactory, albeit with reliability concerns as its accuracy occasionally dips below the $95$\% threshold. This has crucial implications for autonomous healthcare systems, particularly given the risks of false negatives in progressive diseases like cancers. Despite these challenges, our proposed method shows excellent efficiency, with an average runtime of approximately $250$ seconds on a $2$ GB GPU for $2048$ points, demonstrating a correlation between the number of input points and processing time, given constant environment and system configurations. Our technique incorporates pixel proximity in its representation, enabling faster extraction of surrounding and abnormal skin color, a process that occurs later in standard CNNs. This is akin to giving a project manager in software engineering advanced insight into a project's future difficulties who otherwise would be restricted in taking long term calls due to the Sliding Window Planning technique. Just as early knowledge could empower the manager to make better decisions, our network benefits from early information, enhancing its judgement and classification capacity. This results in improved stability in the network's decision-making.
In conclusion,  our system shows promise in medical image analysis with minimal pre-processing. This research has highlighted the importance of data representation, with the potential of point clouds in medical imaging becoming evident. We encourage researchers to challenge data representation biases and remain open to novel forms. Despite promising results, further exploration is warranted. This includes potential improvements via various other surface representations and through advanced architectures like PointNet++, VoxelNet, Point Transformer, or Graph Dynamic CNN. Other areas include testing architectures that accommodate a larger number of points, experimenting with alternative pooling strategies, enhancing results through advanced pre-processing techniques, and further exploration of data representation. 

\newpage
\setcitestyle{numbers}
\bibliographystyle{unsrtnat}
\bibliography{ref}

\newpage

\section{Supplementary}

This supplementary document is intended to provide additional details and references pertaining to our methodology.

\subsection{Mathematical Underpinnings}

Here we will describe some key mathematical results which form the foundation of our methodology. 
A digital image, with its distinct pixel values and positional characteristics, can be considered a topological space. In the context of our work, an image $\mathcal{I}$ is considered as a function $f: \mathbb{Z}^+_m \times \mathbb{Z}^+_n \rightarrow \mathbb{R}^k$, where the input set represents the pixel positions. Considering a digital image in this way allows us to utilize concepts and tools from topology and differential geometry for image analysis. For instance, when defining a hypersurface from an image, we're leveraging the image's inherent topological properties to construct an abstract geometric space that encapsulates the image's structural information.
When transitioning from structured to unstructured data, the representation scheme for the point cloud data is crucial. We use a Cartesian coordinate system, which simplifies the computation of distances between points in space. This approach aids in the construction of the hypersurface $\mathcal{H}$ and adjacency tensor $A$.
The construction of the hypersurface using a coordinate system can be justified by Nash embedding theorem which states that any Riemannian manifold can be embedded isometrically into some Euclidean space  \cite{lee2000introduction}. This theorem supports our approach of mapping a 2D image, whose dense point cloud can be considered a manifold, into a higher-dimensional Euclidean space.\\

The Fourier transform is a powerful tool in image processing, as it allows us to transition from the spatial domain to the frequency domain. This transformation provides information about the frequency content of the image. The Parseval's theorem  guarantees that the energy of the signal is conserved in the frequency domain, making it possible to analyze the image based on its frequency content without losing crucial information.

\textbf{Fourier Transform:} The Fourier Transform is a mathematical tool that is used to transform functions from the time domain (or spatial domain) into the frequency domain. Its existence and properties are guaranteed by the following theorem:\\

\begin{theorem}
     For any function $f(t)$ that is integrable (in the Lebesgue sense) on the real line $\mathbb{R}$, the Fourier Transform $F(\omega)$ exists and is given by the formula:
    \begin{equation}
    F(\omega) = \int_{-\infty}^{\infty} f(t) e^{-2\pi i \omega t} dt
    \end{equation}
    where $i$ is the imaginary unit, and $\omega$ is the frequency variable. 
\end{theorem}
The inverse Fourier Transform exists under certain conditions  and is given by:
    \begin{equation}
    f(t) = \int_{-\infty}^{\infty} F(\omega) e^{2\pi i \omega t} d\omega
    \end{equation}

The computation of curvature at each point of the sparse point cloud is crucial for the formation of the dense point cloud. Curvature is a measure that denotes how much a curve deviates from being a straight line (or a surface from being a plane). We use Gaussian curvature as a measure in our methodology when $k = 3$. Gaussian curvature is an intrinsic measure of curvature, meaning it does not depend on the particular embedding of the surface in space, but only on the way distances are measured along the surface. This property is ensured by Gauss's Theorema Egregium. It is a fundamental result in differential geometry proven by Carl Friedrich Gauss that concerns the curvature of surfaces. The theorem states that Gaussian curvature of a surface is an intrinsic property of the surface, meaning it does not change if the surface is isometrically deformed (bent without stretching).\\

\begin{theorem}
If $f: M \rightarrow \mathbb{R}^3$ is an isometric immersion and $\kappa$ is the Gaussian curvature of $M$, then for any isometric immersion $f': M \rightarrow \mathbb{R}^3$ we have that the Gaussian curvature of the immersed surface $f'(M)$ is also $\kappa$.
\end{theorem}

In higher dimensions, instead of the Gaussian curvature, we have sectional curvature. For an n-dimensional Riemannian manifold $M$, the sectional curvature $K$ at a point $p$ is a function of a 2-dimensional linear subspace $P$ of the tangent space $T_pM$ at $p$.

An analogous statement in arbitrary dimensions would be:

\begin{theorem}
    If $f: M \rightarrow \mathbb{R}^n$ is an isometric immersion and $\kappa$ is the sectional curvature of $M$, then for any isometric immersion $f': M \rightarrow \mathbb{R}^n$ we have that the sectional curvature of the immersed manifold $f'(M)$ is also $\kappa$.
\end{theorem} 

\textbf{The Hessian Matrix:} The Hessian matrix is a square matrix of second-order partial derivatives of a scalar-valued function. The Hessian matrix and its eigenvalues play a crucial role in determining the local maxima, minima, or saddle points of a function. \\

 The Gauss-Bonnet Theorem is a fundamental theorem that relates the Gaussian curvature of a surface to its topology:

\begin{theorem}
    \textbf{(Gauss-Bonnet Theorem)} For a compact surface $\mathcal{S}$, the integral of the Gaussian curvature $K$ over $\mathcal{S}$ is equal to $2\pi$ times the Euler characteristic of $\mathcal{S}$:
    \begin{equation}
    \int_{\mathcal{S}} K \, dA = 2\pi \chi_{\mathcal{S}}
    \end{equation}
\end{theorem}

This theorem provides a critical link between local geometric properties (curvature) and global topological properties (Euler characteristic) of a surface.\\

An important step in our methodology is the estimation of local surface curvature. This is achieved by fitting a local surface to a point and its neighbors in the sparse point cloud. In three dimensions, one common approach for this purpose is to fit a quadratic surface to these points, which is essentially a local Taylor approximation. In the context of a grayscale image, the image intensity function $I(x, y)$ can be viewed as a surface in $\mathbb{R}^3$, and its Gaussian curvature at each point can be calculated as in equation (7),
where $\nabla I_{\psi(\mathcal{S})_{i,j}}$ is the gradient of the image intensity at point $(i, j)$, and $H$ is the Hessian matrix defined as in equation (8).
In this equation, the Hessian matrix $H$ represents the second derivatives of the image intensity, which capture the local geometric information about the image surface.\\

\begin{theorem} \textbf{(Local Surface Curvature Estimation)}
    Let $S: \mathbb{R}^2 \rightarrow \mathbb{R}$ be a smooth function representing the surface, with $S(0,0)$ being the point where we want to estimate the curvature. The quadratic approximation of $S$ around $(0,0)$ is given by:
\begin{equation}
S(x, y) \approx S(0,0) + xS_x + yS_y + x^2S_{xx} + 2xyS_{xy} + y^2S_{yy},
\end{equation}
where $S_x$, $S_y$, $S_{xx}$, $S_{xy}$, and $S_{yy}$ are partial derivatives of $S$. 
\end{theorem}
From this approximation, we can calculate the curvature of the surface at $(0,0)$ by using the eigenvalues of the Hessian matrix.\\

The formation of the probability distribution based on the curvature values leans on the principle of importance sampling. In this context, points with higher curvature are given more importance (higher probability), allowing for a denser sampling in those areas during the creation of the dense point cloud. This technique guarantees that more detailed regions are adequately represented in the transformed representation. The Monte Carlo sampling method we utilize is a powerful statistical technique that allows for the estimation of an unknown quantity based on the principles of inferential statistics. The central limit theorem  justifies the usage of Monte Carlo methods by implying that the sum of a large number of independent and identically distributed (i.i.d.) random variables will be approximately normally distributed, irrespective of the shape of the original distribution.\\

\newpage

 Given a probability distribution $P$, the Monte Carlo algorithm generates samples by following these steps:\\
\begin{itemize}
    \item Generate a random number $r$ from a uniform distribution in the interval $[0, 1]$.
    \item Find the smallest $i$ such that $P(1) + P(2) + ... + P(i) \geq r$.

    \item The sample is $i$.
\end{itemize}

This sampling method is used in our work to generate a dense point cloud from a sparse one, based on a probability distribution that reflects the geometric characteristics of the sparse point cloud. This ensures that the most crucial information from the original image is preserved in the dense point cloud.

 \textbf{Nyquist-Shannon Sampling Theorem} This is one of the fundamental concepts in signal processing and data representation that has been implicitly used in our methodology. It suggests that a signal can be perfectly reconstructed from its samples if the sampling frequency is greater than twice the highest frequency present in the signal. In the context of our work, this theorem is crucial when we convert a $2D$ image into a point cloud and then sample this point cloud to create a denser representation. The sampling frequency should be chosen carefully to ensure that the denser point cloud preserves the important details of the original image.
\begin{theorem} 
   If a function $f(t)$ contains no frequencies higher than $B$ hertz, it is completely determined by giving its ordinates at a series of points spaced $1/(2B)$ seconds apart. 
\end{theorem}

\subsection{Result Verification and Validity}

Given the inherent randomness of our methodology, due to the stochastic nature of dense point cloud generation, it is crucial to verify the results' validity and repeatability. We used the following strategies to accomplish this:

\begin{itemize}

\item \textbf{Multiple Runs:} We performed multiple runs of the methodology, each with a different random seed, verifying that the outputs are consistent across runs. This ensures that the results are robust to the inherent randomness of the methodology.

\item \textbf{Comparative Analysis:} We compared the transformed images with the original images and evaluated how well the important features of the images are preserved. This can be quantitatively measured using metrics like Structural Similarity Index (SSIM) or Peak Signal-to-Noise Ratio (PSNR).

\item \textbf{Testing on Various Image Types:} We applied our methodology to various types of images (e.g., skin medical images) and verified that it performs well across these different types. This ensures that our methodology is generalizable and not tied to a specific type of image.

\item \textbf{Statistical Analysis:} We  performed a statistical analysis of the generated point clouds (e.g., studying the distribution of the points, the average distance between points, etc.) to understand the properties of the generated data better. 

\end{itemize}

\subsection{License}
We will make our code, dataset, and models accessible to the public. We have utilised established licenses within the community, and will provide corresponding links to these licenses for the datasets, codes, and models involved in our project.


\end{document}